\documentclass[letterpaper, 10 pt, conference]{ieeeconf}

\IEEEoverridecommandlockouts
\usepackage{amsmath,amsfonts}
\usepackage{array}
\usepackage{url}
\usepackage{graphicx}
\usepackage{cite}
\usepackage{float}
\usepackage[hidelinks]{hyperref}

\ifdefined\pdfsuppressptexinfo
\fi

\begin{document}

\title{\LARGE \bf
MMUSV-Sim: A Maritime Simulation Platform for Multi-USV Cooperative Perception
}

\author{Ziao Li$^{1}$, Jianxiong Ye$^{1}$, Biao Tang$^{1}$, Leping Zhang$^{2}$, Kun Zuo$^{1}$,\\
Siyu Huang$^{1}$, Mengtang Li$^{1}$, Deyu Meng$^{3}$, and Chenqiang Gao$^{1,*}$%
\thanks{$^{1}$School of Intelligent Systems Engineering, Sun Yat-sen University, Shenzhen 518107, Guangdong, China. E-mails: \mbox{lizao@mail2.sysu.edu.cn} (Ziao Li); \mbox{gaochq6@mail.sysu.edu.cn} (Chenqiang Gao). $^{*}$Corresponding author: Chenqiang Gao.}%
\thanks{$^{2}$School of Intelligent Imagery Engineering, Beijing Film Academy, Beijing, China.}%
\thanks{$^{3}$School of Mathematics and Statistics, Xi'an Jiaotong University, Xi'an, China.}%
}

\maketitle
\thispagestyle{empty}
\pagestyle{empty}

\begin{abstract}
Perception from a single unmanned surface vehicle (USV) is limited by sensing range, field of view, and occlusions caused by islands, port infrastructure, and nearby vessels. Multi-USV cooperative perception can alleviate these limitations by integrating complementary observations from distributed viewpoints. However, progress is constrained by the high cost of coordinated sea trials, the scarcity of multi-USV datasets, and the lack of integrated maritime simulation and evaluation tools. We present MMUSV-Sim, a perception-oriented maritime simulation and data-generation platform built on Unreal Engine 5 and Project AirSim. It supports configurable maritime scenarios, synchronized multimodal acquisition across multiple USVs, and consistent 3D annotations in a shared world frame. Using MMUSV-Sim, we construct MarU2U-200m, a synchronized multi-USV RGB--LiDAR dataset with 341 sequences and 14,385 scene frames. We establish a cooperative bird's-eye-view (BEV) vessel detection benchmark with LiDAR and RGB--LiDAR inputs. Experimental results show that Early Fusion improves LiDAR detection AP@0.5 from 64.26 to 81.46 compared with the single-USV baseline. RGB features further improve intermediate fusion, increasing AP@0.5 by up to 4.55 percentage points. LiDAR and RGB transfer experiments also show the value of the generated data as a pretraining source for real maritime detection. The project are available at \url{https://github.com/lizaowo/MMUSV-Sim}.
\end{abstract}

\section{Introduction}
\begin{figure}[t]
    \centering
    \includegraphics[width=0.9\linewidth]{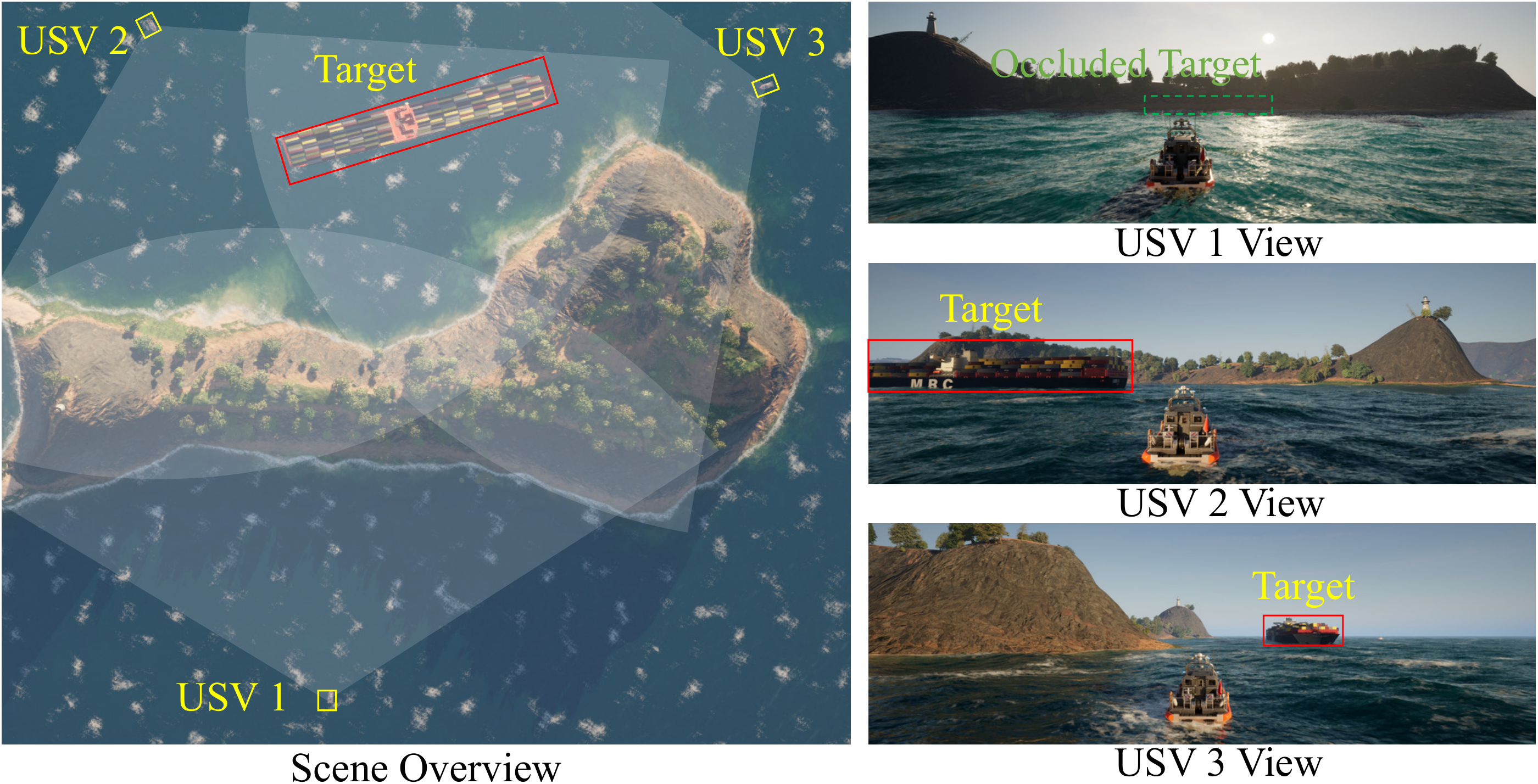} 
    \caption{Illustration of a cooperative perception scenario in MMUSV-Sim. The target vessel is occluded from USV 1 by the island, while USV 2 and USV 3 observe it from complementary viewpoints. Their observations can be expressed in a shared BEV frame for cooperative perception.}
    \label{fig:teaser}
\end{figure}

Unmanned surface vehicles (USVs) are increasingly deployed in maritime operations such as port inspection, waterway monitoring, and autonomous navigation. Reliable detection of surrounding vessels~\cite{KimRAL2023,ClunieICRA2021} is essential for safe operation in complex maritime environments. However, onboard perception from a single USV is constrained by sensing range, limited field of view, and occlusions caused by islands, shoreline structures, port infrastructure, and nearby vessels. Cooperative perception combines observations from multiple USVs in a shared reference frame, using complementary viewpoints to mitigate these limitations, as illustrated in Fig.~\ref{fig:teaser}. Progress in maritime cooperative perception remains limited by the scarcity of multi-USV datasets.

Real-world multi-USV data collection for cooperative perception is difficult and resource-intensive~\cite{Han2026}. Such campaigns require multiple sensor-equipped USVs and carefully coordinated sea trials. Weather and surrounding traffic are difficult to control. Safety constraints and limited test time also restrict the coverage of vessel layouts, viewpoints, target ranges, and occlusion conditions. Beyond acquisition, constructing a cooperative-perception dataset requires synchronizing observations, registering them in a common coordinate frame, and annotating targets consistently across agents. Most public real-world maritime datasets target single-platform perception. Vision-oriented datasets such as MaSTr1325~\cite{MaSTr1325} and KOLOMVERSE~\cite{KOLOMVERSE} mainly provide image or video data with 2D annotations, while multimodal datasets such as SeePerSea~\cite{SeePerSea} and USVTrack~\cite{USVTrack} provide combinations of RGB, LiDAR, and radar observations. Consequently, these datasets provide limited support for training and evaluating synchronized multi-USV cooperative perception.

\begin{table}[t]
\caption{Comparison of representative maritime simulation resources and their perception data-generation capabilities. Coop. denotes dedicated multi-USV perception-data acquisition.}
\label{tab:relatedwork}
\centering
\footnotesize
\setlength{\tabcolsep}{2pt}
\renewcommand{\arraystretch}{1.08}
\begin{tabular*}{\columnwidth}{@{\extracolsep{\fill}}>{\raggedright\arraybackslash}p{0.20\columnwidth}>{\raggedright\arraybackslash}p{0.24\columnwidth}>{\raggedright\arraybackslash}p{0.42\columnwidth}>{\centering\arraybackslash}p{0.08\columnwidth}@{}}
\hline
Resource & Engine & Sensors & Coop. \\
\hline
Virtual RobotX~\cite{VRX} & Gazebo & RGB, LiDAR & No \\
Stonefish~\cite{Stonefish} & Bullet/OpenGL & RGB, depth, sonar, thermal & No \\
MARUS~\cite{MARUS} & Unity & RGB, LiDAR, sonar & No \\
ASVSim~\cite{ASVSim} & UE5 / Cosys-AirSim & RGB, depth, semantic, LiDAR, radar & No \\
S2S-Sim~\cite{S2SSim} & Unity & LiDAR & Yes \\
\hline
\textbf{MMUSV-Sim} & \textbf{UE5 / Project AirSim} & \textbf{RGB, depth, semantic, LiDAR, radar} & \textbf{Yes} \\
\hline
\end{tabular*}
\end{table}

\begin{figure*}[t]
    \centering
    \includegraphics[width=0.8\linewidth]{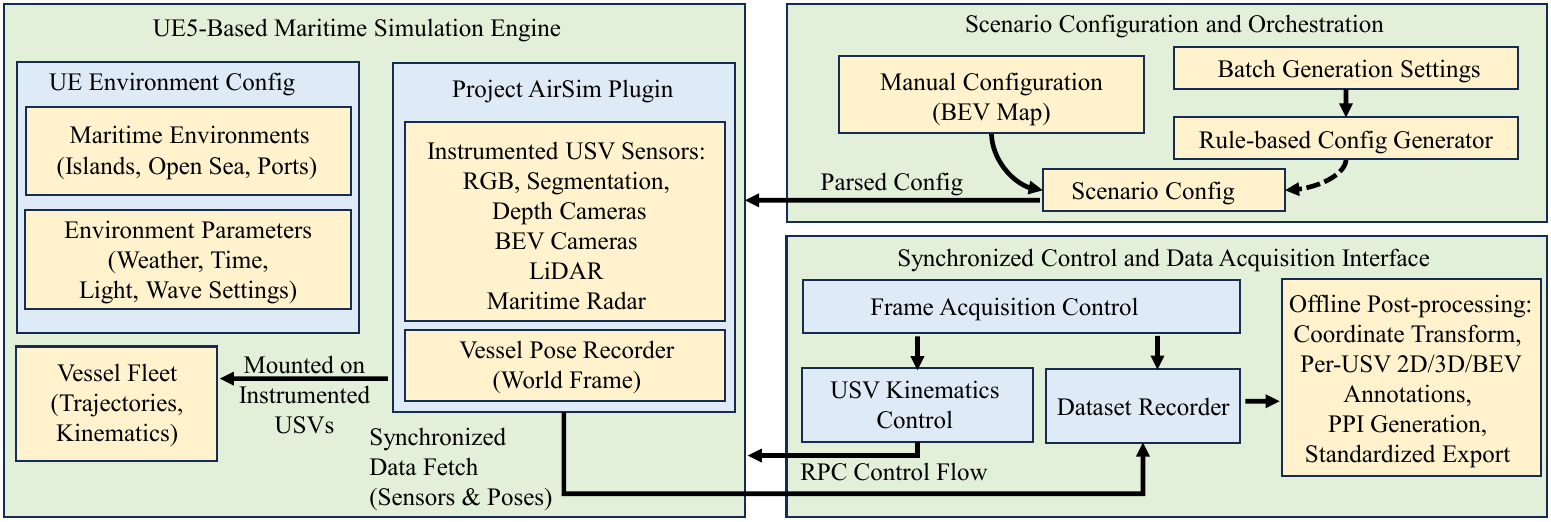} 
    \caption{Overall architecture of MMUSV-Sim. Manual and batch-generated scenario configurations instantiate the UE5 and Project AirSim environment. The control and acquisition interface coordinates vessel motion, multimodal recording while scene evolution is paused, vessel pose recording, and offline generation of per-agent annotations and structured exports.}
    \label{fig:architecture}
\end{figure*}

Simulation is an important way to complement costly real-world data collection. In autonomous driving and aerial robotics, resources such as OPV2V~\cite{OPV2V}, V2X-Sim~\cite{V2XSim}, U2USim~\cite{U2USim}, and Griffin~\cite{griffin} have enabled controllable multi-agent data generation and standardized cooperative-perception evaluation. However, these resources are tailored to road or aerial settings and do not directly represent maritime characteristics such as open-water layouts, island and port occlusions, large variations in vessel scale, and wave-related sensor attitude changes. Marine robotics simulators such as Virtual RobotX~\cite{VRX}, Stonefish~\cite{Stonefish}, MARUS~\cite{MARUS}, and ASVSim~\cite{ASVSim} provide vehicle dynamics, environmental modeling, sensor simulation, and, in several cases, automatic annotation. As summarized in Table~\ref{tab:relatedwork}, they are primarily designed for general marine-robotics simulation rather than dedicated multimodal data-generation pipelines for multi-USV perception. S2S-Sim~\cite{S2SSim} provides a cooperative 3D ship-detection dataset and benchmark, but its released observations are limited to LiDAR. \textit{Overall, existing resources do not yet provide an integrated maritime pipeline for multimodal multi-USV data generation, synchronized acquisition, and unified world-frame annotation.}

To address these limitations, we present MMUSV-Sim, a maritime simulation and data generation platform for perception research. Built on Unreal Engine 5 (UE5) and Project AirSim~\cite{airsim}, the platform supports configurable maritime scenarios. It produces synchronized multimodal observations from multiple USVs, together with consistent 3D annotations in a shared world coordinate frame. Using MMUSV-Sim, we construct MarU2U-200m, a synchronized multi-USV RGB and LiDAR dataset covering island, open sea, and port environments. Bird's-eye-view (BEV) detection is widely used for cooperative perception in autonomous driving~\cite{OPV2V,V2XSim}. Motivated by this approach, we establish a cooperative BEV vessel detection benchmark with LiDAR and RGB--LiDAR inputs as a representative application of MMUSV-Sim. The benchmark evaluates the benefits of shared observations across USVs and additional visual information through comparisons of cooperative fusion methods. It further examines communication cost, vessel scale, sensing range, pose noise, and temporal delay. Transfer experiments using LiDAR and RGB assess whether the generated data provide useful pretraining for real maritime detection.
The main contributions are summarized as follows:

\begin{itemize}
\item \textbf{Maritime Simulation and Data Generation Platform:}
MMUSV-Sim integrates configurable maritime environments and traffic scenarios, multimodal acquisition across multiple USVs, and annotation generation from platform and vessel states recorded in a shared world frame. The resulting observations and annotations for each USV are organized as structured exports for downstream perception.

\item \textbf{Multi-USV RGB and LiDAR Dataset:}
MarU2U-200m provides synchronized surround-view RGB images, LiDAR observations, recorded USV poses, and vessel annotations across island, open sea, and port environments. These records support standardized training and evaluation of multi-USV cooperative perception.

\item \textbf{Representative Cooperative Benchmark and Transfer Validation:} The cooperative BEV benchmark compares early, intermediate, and late fusion with LiDAR and RGB--LiDAR inputs. It evaluates the benefits of shared observations and visual information, together with communication cost and performance across vessel scales, sensing ranges, pose noise, and temporal delay. Transfer experiments demonstrate the value of MarU2U-200m pretraining for LiDAR and RGB detection on real maritime datasets.
\end{itemize}

\begin{figure*}[t]
\centering
\includegraphics[width=0.7\textwidth]{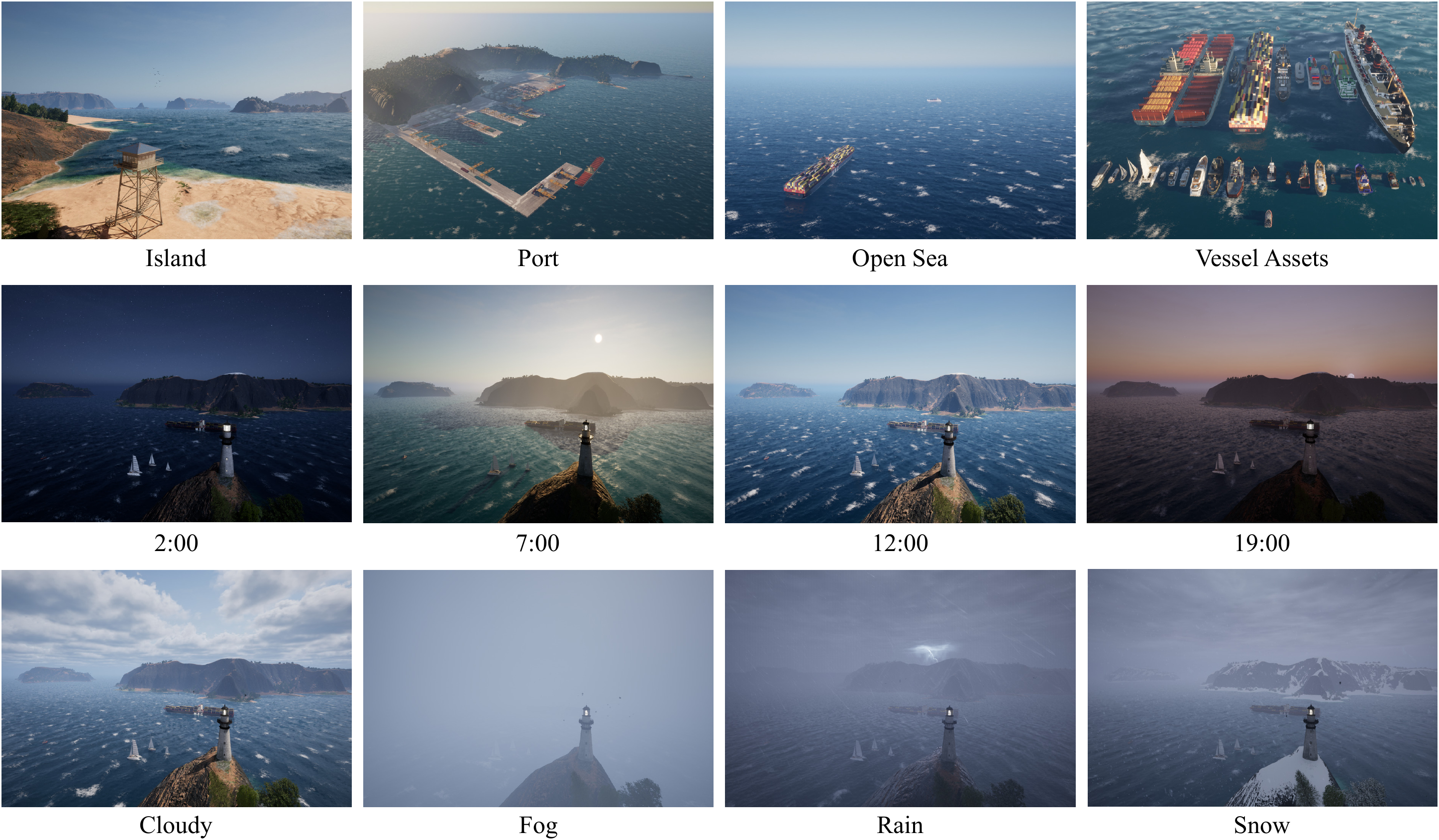}
\caption{Visual diversity and environmental controllability of MMUSV-Sim. Top row: representative island, port, and open sea topologies, together with the vessel asset library. Middle row: time-of-day variations under a fixed scene configuration. Bottom row: weather variations under a fixed scene configuration.}
\label{fig:visual_diversity}
\end{figure*}

\section{MMUSV-Sim System Overview}
As shown in Fig.~\ref{fig:architecture}, MMUSV-Sim consists of three connected components: a UE5-based maritime simulation engine, a scenario configuration and orchestration module, and a multi-USV data acquisition and export interface. The scenario module converts manual BEV inputs or rule-based batch settings into structured configurations that define the environment, vessels, trajectories, and sensors. Manual and batch-generated configurations share the same scene execution, acquisition, and annotation pipeline. The acquisition interface records synchronized multimodal observations and vessel poses. Offline post-processing generates per-agent annotations and structured exports.

\subsection{UE5-Based Maritime Simulation Engine}

MMUSV-Sim adopts UE5 as the simulation engine for maritime scene rendering, multi-vessel instantiation, sensor mounting, and perception-oriented data generation. We construct three representative maritime environments: island, open sea, and port, which provide different spatial layouts and shoreline structures for maritime perception studies. The engine provides parameterized control over time of day, weather, illumination, and wave settings within a common simulation framework, as shown in Fig.~\ref{fig:visual_diversity}. Oceanology provides configurable sea-surface waves.

To represent the substantial variations in maritime target geometry and scale, the platform integrates more than 30 vessel assets, with lengths ranging from approximately 3~m for speedboats to approximately 300~m for cargo vessels. Together with the scenario configuration module, this asset library supports diverse multi-vessel layouts, target ranges, and occlusion conditions for multi-USV cooperative perception. Vessels follow configured spline paths at nominal speeds, with headings determined by the local path direction. Synthetic heave, roll, and pitch variations are superimposed on the route motion, with amplitudes scaled by the wave setting and vessel size. Vessel poses are updated kinematically for perception data generation rather than computed through full vessel hydrodynamics.

Through a customized integration of Project AirSim within UE5, MMUSV-Sim provides instrumented USVs with a configurable multimodal sensor suite, including surround-view RGB cameras, depth cameras, semantic segmentation cameras, top-view cameras for BEV rendering, a spinning LiDAR, and a ray-cast maritime radar. Scenario configurations specify sensor resolution, field of view, mounting pose, update rate, LiDAR channels, and sensing range. The ray-cast radar models configurable geometric visibility and range through sparse range--azimuth--elevation returns for offline plan position indicator (PPI) generation. Each sequence uses fixed sensor-to-USV extrinsics for annotation projection and spatial cross-modal alignment.

\begin{figure*}[t]
\centering
\includegraphics[width=0.8\textwidth]{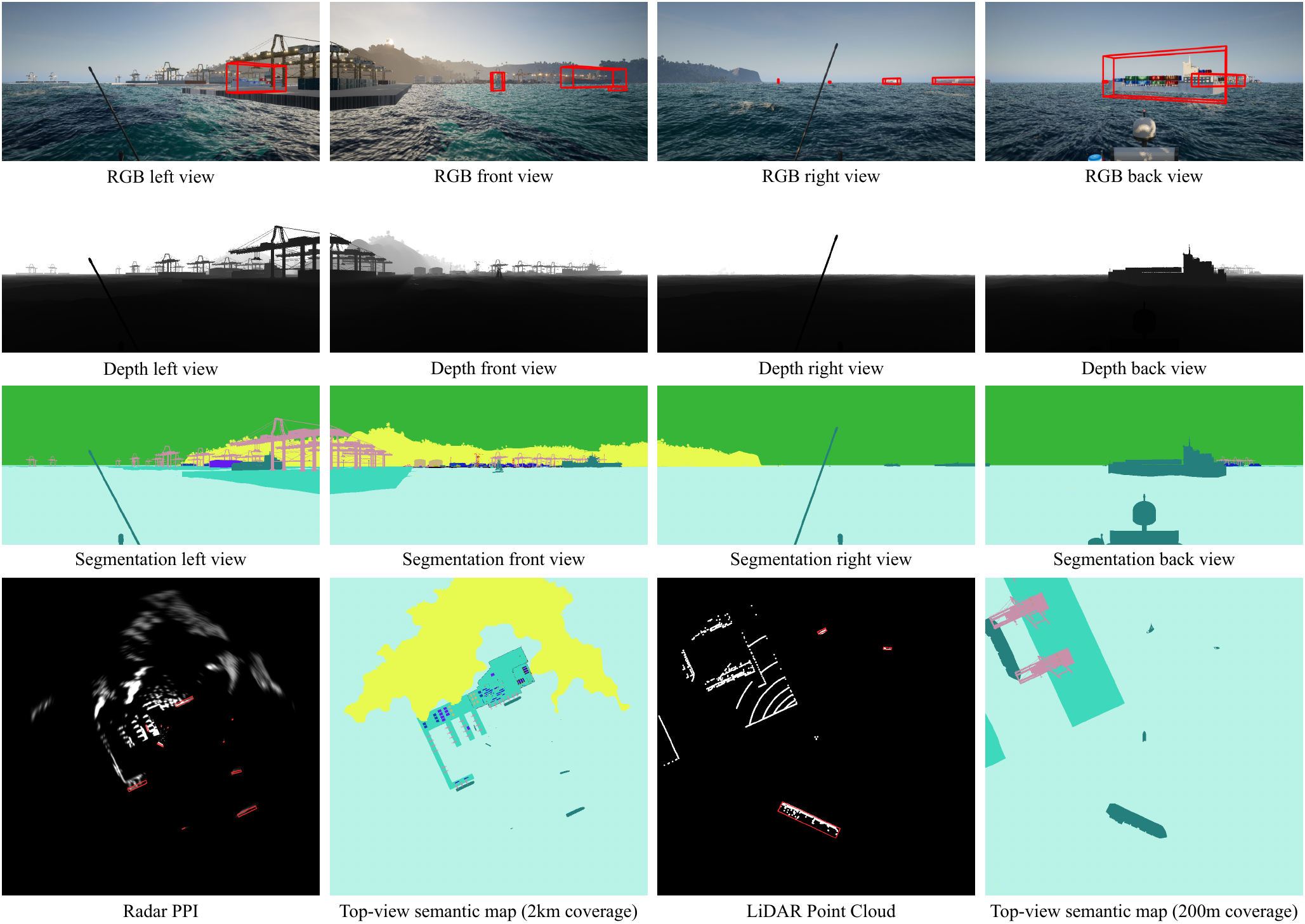}
\caption{Sensor and annotation visualizations generated by MMUSV-Sim, including surround-view RGB images, depth maps, semantic segmentation, radar PPI, top-view semantic maps, and a LiDAR Point Cloud.}
\label{fig:multimodal_outputs}
\end{figure*}

\subsection{Multi-USV Data Acquisition and Export}
The control and data-acquisition interface connects scenario execution with the Project AirSim sensor streams. At each acquisition cycle, the interface pauses scene evolution while keeping the sensor streams active. After the sensor buffers are refreshed for the paused scene state, the interface records the updated camera images, LiDAR point clouds, and radar returns, together with the USV and target-vessel poses. The resulting sensor observations and poses correspond to the same paused scene state and are assigned a common acquisition index.

Target vessel poses are first recorded in the shared world frame. During offline post-processing, the recorded USV poses and configuration-defined sensor extrinsics are used to transform these poses into the corresponding USV body and sensor frames. Using the transformed poses and target vessel geometry, the platform generates camera-projected 2D boxes, LiDAR-frame 3D boxes, and ego-centric BEV boxes. Sparse radar returns are rasterized into PPI representations. The processed records are organized into structured exports for downstream cooperative perception tasks. Representative multimodal outputs produced by this acquisition and post-processing pipeline are shown in Fig.~\ref{fig:multimodal_outputs}.

\section{MarU2U-200m Dataset}

\subsection{Dataset Scope and Sensor Configuration}
Using MMUSV-Sim, we construct \textbf{MarU2U-200m}, a synchronized multi-USV RGB and LiDAR dataset for cooperative vessel perception with a 200~m LiDAR sensing range. Each sequence contains 2--4 instrumented USVs and multiple target vessels. At each acquisition index, observations from all instrumented USVs are associated with the corresponding USV and target vessel poses in the shared world frame, enabling consistent annotations across USVs.

\begin{figure}[t]
    \centering
    \includegraphics[width=\linewidth]{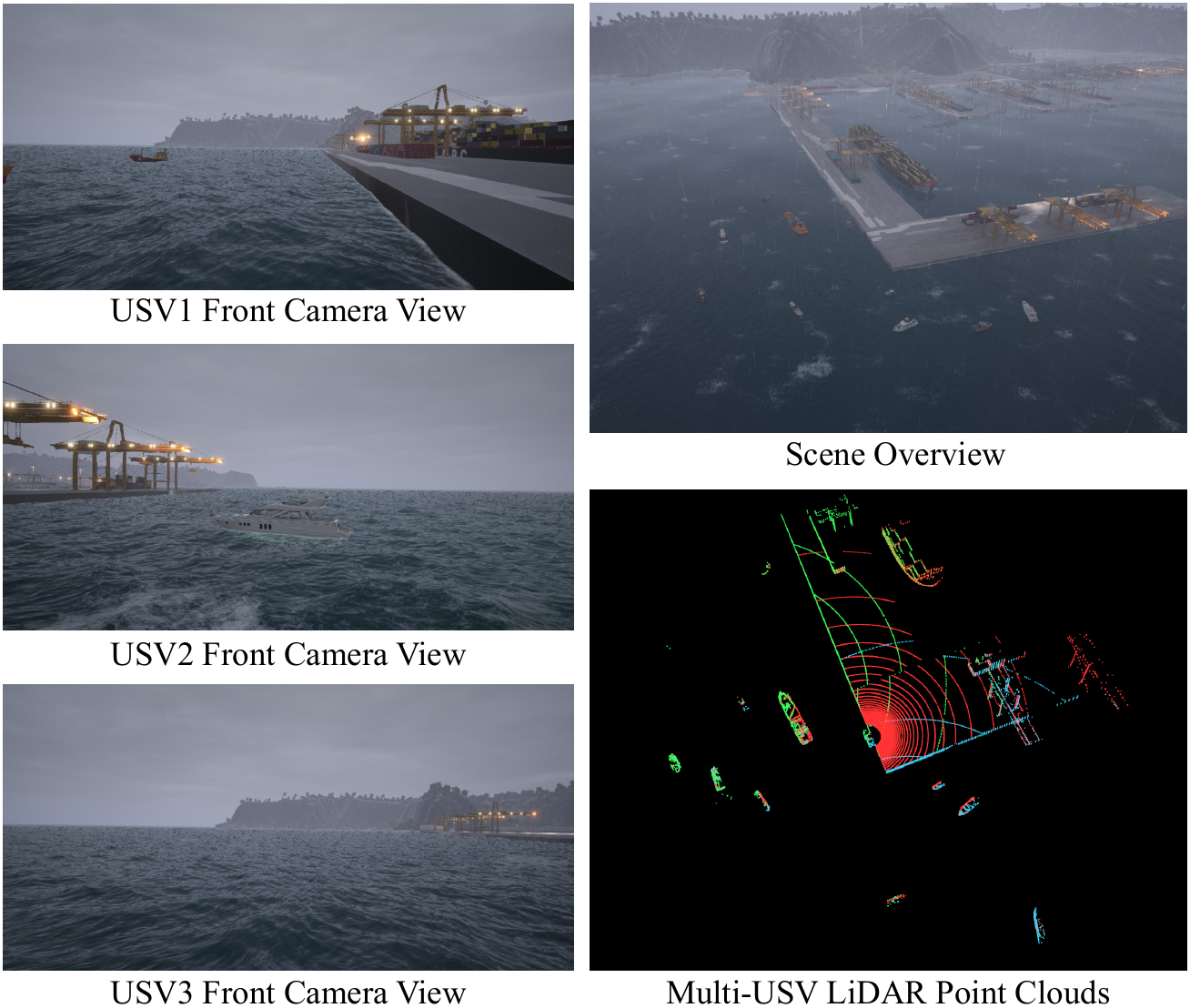}
    \caption{Qualitative visualization of a synchronized multi-USV sample in \textbf{MarU2U-200m}. The figure shows onboard RGB observations, a scene-level rendering for visual reference, and registered multi-USV LiDAR point clouds in the ego LiDAR frame. Red, green, and blue points correspond to USV 1, USV 2, and USV 3, respectively.}
    \label{fig:multi_usv_observation}
\end{figure}

MarU2U-200m contains 341 multi-USV sequences and 14,385 synchronized scene frames. Each scene frame groups observations from all instrumented USVs at one acquisition index, with each USV contributing one per-agent record. The dataset therefore contains 33,525 per-agent records, comprising 33,525 LiDAR sweeps and 134,100 RGB images. The sensor configuration of each instrumented USV is summarized in Table~\ref{tab:sensor_config}, and a representative multi-USV sample is shown in Fig.~\ref{fig:multi_usv_observation}.

\begin{table}[t]
\caption{Sensor Configuration of Each Instrumented USV in MarU2U-200m.}
\label{tab:sensor_config}
\centering
\footnotesize
\setlength{\tabcolsep}{2pt}
\begin{tabular*}{\columnwidth}{@{\extracolsep{\fill}}>{\raggedright\arraybackslash}p{0.18\columnwidth}>{\raggedright\arraybackslash}p{0.54\columnwidth}>{\centering\arraybackslash}p{0.12\columnwidth}>{\centering\arraybackslash}p{0.09\columnwidth}@{}}
\hline
Sensor & Configuration & Capture rate & Output \\
\hline
RGB cameras & Four-view surround cameras, $1024 \times 512$ resolution, $90^\circ$ FOV each & 5~Hz & PNG \\
LiDAR & 64 channels, 360$^\circ$ horizontal FOV, $[-20^\circ,10^\circ]$ vertical FOV, 200~m range & 5~Hz & PCD \\
\hline
\end{tabular*}
\end{table}

\subsection{Coordinate Frames and Annotation Protocol}
For each instrumented USV, target vessel poses and 3D bounding boxes are transformed from the shared world frame into the corresponding LiDAR and camera frames using the recorded USV pose and sensor extrinsics. Consequently, annotations across USVs and sensing modalities are derived from the same target geometry. For LiDAR-based detection, each target vessel is represented by a 3D bounding box $B=(x,y,z,l,w,h,\theta)$ in the corresponding USV LiDAR frame, where $(x,y,z)$ denotes the box center, $(l,w,h)$ denote its length, width, and height, and $\theta$ denotes the box yaw angle in the LiDAR frame. For BEV detection, the boxes retain the vessel dimensions and yaw, while roll and pitch are set to zero. BEV boxes and camera-projected 2D boxes are derived from the same target vessel geometry.

MarU2U-200m provides two annotation views. The \emph{LiDAR-visible view} retains a target vessel when at least one LiDAR point from the corresponding USV lies inside its exported 3D box, and is used for training. The \emph{complete annotation view} includes all target vessels within the 200~m sensing range after transformation into the ego LiDAR frame, including those not observed by the ego LiDAR. We use this complete view for evaluation to measure whether cooperative observations can recover targets unavailable from the ego viewpoint.

\subsection{Dataset Statistics and Split}

\begin{table}[t]
\caption{Sequence-Level Split by Maritime Topology of MarU2U-200m. Each entry reports sequences / scene frames.}
\label{tab:data_split}
\centering
\footnotesize
\setlength{\tabcolsep}{4pt}
\begin{tabular*}{\columnwidth}{@{\extracolsep{\fill}}lcccc@{}}
\hline
Split & Island & Open Sea & Port & Total \\
\hline
Training & 108 / 4,085 & 31 / 1,358 & 100 / 4,058 & 239 / 9,501 \\
Validation & 10 / 376 & 11 / 548 & 13 / 784 & 34 / 1,708 \\
Testing & 23 / 1,029 & 18 / 788 & 27 / 1,359 & 68 / 3,176 \\
\hline
Total & 141 / 5,490 & 60 / 2,694 & 140 / 6,201 & 341 / 14,385 \\
\hline
\end{tabular*}
\end{table}

MarU2U-200m is split by sequence to avoid data leakage across the training, validation, and test sets. As shown in Table~\ref{tab:data_split}, the resulting split contains 239 training sequences, 34 validation sequences, and 68 testing sequences across island, open sea, and port environments. The dataset contains 34 target vessel models ranging in length from 4.91~m to 314.89~m, together with 271 two-USV, 64 three-USV, and 6 four-USV sequences. Across all per-agent records, the complete annotation view contains 88,007 3D vessel boxes within the 200~m LiDAR sensing range, including 24,542 boxes at distances of 150--200~m.

\section{Cooperative BEV Benchmark}
We establish a cooperative BEV vessel detection benchmark on MarU2U-200m as a representative application of MMUSV-Sim. The benchmark compares fusion methods using LiDAR and RGB--LiDAR inputs to evaluate the benefits of cooperation and additional visual information. Communication cost is reported for both input settings. We use LiDAR for vessel scale, sensing range, and robustness analyses, focusing on geometric observations and their spatial and temporal alignment.

\subsection{Benchmark Protocol and Experimental Setup}
The task is to predict oriented vessel bounding boxes in the ego USV's BEV frame using local and shared observations. Both input settings use the same sequence split and evaluation protocol. Each sequence specifies a fixed ego USV, and this assignment is shared across all methods. Models are trained with the LiDAR-visible annotation view and evaluated with the complete annotation view within the 200~m evaluation range. During training, early and intermediate fusion use the union of LiDAR-visible annotations from the participating USVs, transformed into the ego frame and deduplicated by target ID. We report BEV average precision (AP) at IoU thresholds of 0.3, 0.5, and 0.7.

The LiDAR setting uses XYZI point clouds from the participating USVs. The detector uses a PointPillars encoder~\cite{pointpillar} and a CenterPoint detection head~\cite{CenterPoint}, with a horizontal pillar size of $0.4 \times 0.4$~m. The RGB--LiDAR setting extends this framework with a MobileNetV3-Small~\cite{MobileNetV3} image encoder initialized with ImageNet-pretrained~\cite{imagenet} weights. It extracts features from four surround-view images resized to $512 \times 256$ pixels. A projection layer maps the image features to 16 channels. LiDAR points are projected into the images using the camera intrinsics and sensor extrinsics. The sampled visual features are concatenated with the XYZI point features before pillar encoding. The detection network and the new projection layer are randomly initialized. All models are trained for 30 epochs using Adam. Checkpoints are selected by the highest validation AP@0.5 among epochs 21--30. Late Fusion reuses the corresponding Single USV detector.

Using the OpenCOOD framework from OPV2V~\cite{OPV2V}, we compare a Single USV baseline and three representative cooperative fusion paradigms. Single USV uses only the ego observation. Early Fusion transforms neighboring point clouds into the ego frame and aggregates them before pillar encoding. Late Fusion transforms independently predicted boxes into the ego frame and merges them using non-maximum suppression. Intermediate Fusion combines aligned BEV features, with F-Cooper~\cite{FCooper}, AttFusion~\cite{OPV2V}, and Where2comm~\cite{Where2comm} evaluated under both input settings. Communication cost measures the average FP32 payload transmitted from neighboring USVs to the ego per scene frame, excluding auxiliary metadata and network protocol overhead. For RGB--LiDAR Early Fusion, communication is estimated assuming local image feature extraction, with XYZI and 16 visual features transmitted per point.

\subsection{Cooperative Perception Results}
\begin{figure*}[t]
    \centering
    \includegraphics[width=0.7\textwidth]{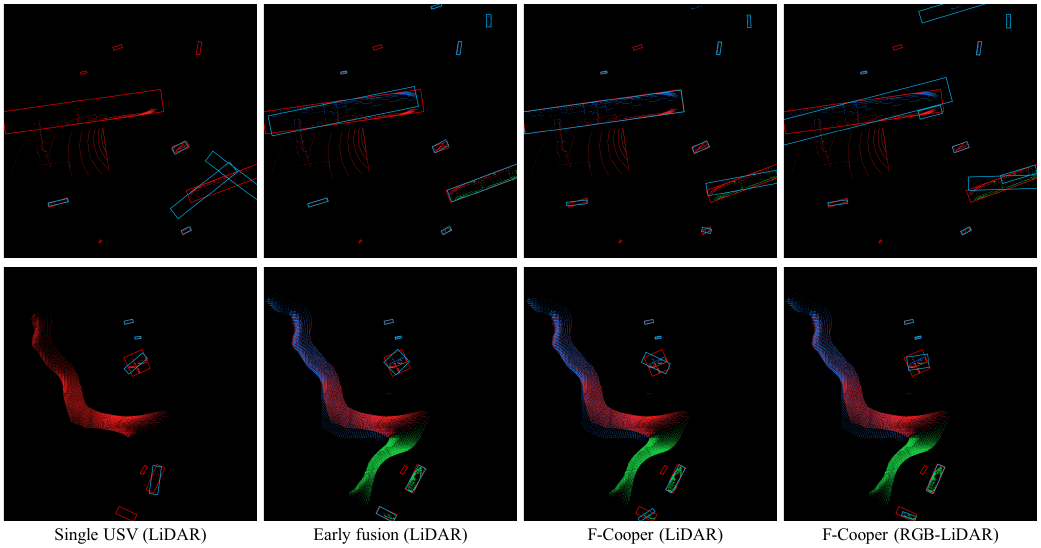}
    \caption{Qualitative comparison of cooperative BEV vessel detection on MarU2U-200m in port (top) and island (bottom) scenes. Red boxes denote ground truth and cyan boxes denote predictions. Point colors indicate observations from different USVs.}
    \label{fig:qualitative_detection}
\end{figure*}

Table~\ref{tab:cooperative_results} shows that most cooperative methods outperform Single USV with LiDAR input. Early Fusion achieves the highest AP at all three IoU thresholds, suggesting that combining partial observations before feature encoding provides useful geometric evidence for vessel detection. Figure~\ref{fig:qualitative_detection} presents qualitative comparisons in port and island scenes. Late Fusion also performs competitively and has the lowest communication cost among cooperative methods, transmitting only predicted boxes. Although raw point sharing is commonly considered costly in autonomous driving~\cite{OPV2V}, open-water regions in our maritime scenes contribute few background returns. Early Fusion therefore transmits only 0.101 MB per frame, compared with 93.441 MB for fixed-size dense BEV features. This makes raw point sharing an efficient option for the sparse maritime observations in this benchmark.

With RGB--LiDAR inputs, Early Fusion remains the best-performing method at all three IoU thresholds, with performance close to its LiDAR-only counterpart. F-Cooper, AttFusion, and Where2comm improve at all three thresholds, gaining 4.55, 2.95, and 0.87 percentage points at AP@0.5, respectively. Single USV and Late Fusion also improve at AP@0.7, while their AP@0.5 decreases slightly. These results show that synchronized visual features can complement LiDAR geometry and improve cooperative vessel detection, with consistent gains across the evaluated intermediate fusion methods. Adding visual features increases the Early Fusion payload to 0.506 MB per frame, while dense intermediate fusion retains the same communication cost.

\begin{table*}[t]
\caption{Cooperative BEV vessel detection results on the MarU2U-200m test set. AP is reported as a percentage, and communication cost is the average non-ego-to-ego payload in MB per frame.}
\label{tab:cooperative_results}
\label{tab:rgb_lidar_results}
\centering
\footnotesize
\setlength{\tabcolsep}{4pt}
\renewcommand{\arraystretch}{1.05}
\begin{tabular*}{\textwidth}{@{\extracolsep{\fill}}llcccccccc@{}}
\hline
& & \multicolumn{4}{c}{LiDAR} & \multicolumn{4}{c}{RGB--LiDAR} \\
\cline{3-6} \cline{7-10}
Method & Fusion & AP@0.3 & AP@0.5 & AP@0.7 & Comm. & AP@0.3 & AP@0.5 & AP@0.7 & Comm. \\
\hline
Single USV & None & 71.58 & 64.26 & 45.93 & 0 & 70.87 & 63.35 & 48.30 & 0 \\
Late Fusion & Late & 83.45 & 77.14 & 60.59 & $1.93 \times 10^{-4}$ & 83.52 & 76.96 & 61.89 & $1.90 \times 10^{-4}$ \\
Early Fusion & Early & \textbf{85.23} & \textbf{81.46} & \textbf{70.06} & 0.101 & \textbf{85.80} & \textbf{81.98} & \textbf{69.72} & 0.506 \\
\hline
F-Cooper~\cite{FCooper} & Intermediate & 78.14 & 69.00 & 51.73 & 93.441 & 80.84 & 73.55 & 54.89 & 93.441 \\
AttFusion~\cite{OPV2V} & Intermediate & 79.56 & 71.14 & 54.15 & 93.441 & 80.56 & 74.09 & 57.19 & 93.441 \\
Where2comm~\cite{Where2comm} & Intermediate & 81.37 & 73.88 & 56.30 & 0.514 & 81.68 & 74.75 & 58.20 & 0.511 \\
\hline
\end{tabular*}
\end{table*}

\subsection{Analysis by Vessel Scale and Range}

\begin{table}[t]
\caption{Scale-wise LiDAR BEV vessel detection on MarU2U-200m. Groups use vessel length; AP@0.5 is reported as a percentage.}
\label{tab:scale_wise_results}
\centering
\footnotesize
\setlength{\tabcolsep}{2.2pt}
\begin{tabular*}{\columnwidth}{@{\extracolsep{\fill}}lcccc@{}}
\hline
& \multicolumn{4}{c}{Vessel Length (m)} \\
\cline{2-5}
Method & $<10$ & 10--30 & 30--100 & $\geq100$ \\
\hline
GT boxes & 3,146 & 9,751 & 4,690 & 1,367 \\
\hline
Single USV & 45.35 & 69.83 & 61.63 & 30.51 \\
Late Fusion & 62.20 & 80.00 & 76.95 & 41.02 \\
Early Fusion & \textbf{64.48} & \textbf{83.12} & \textbf{83.55} & \textbf{54.88} \\
\hline
F-Cooper & 53.12 & 74.63 & 62.00 & 41.74 \\
AttFusion & 58.73 & 77.97 & 60.95 & 38.11 \\
Where2comm & 55.69 & 77.95 & 73.10 & 44.32 \\
\hline
\end{tabular*}
\end{table}

\begin{table}[t]
\caption{Range-wise LiDAR BEV vessel detection on MarU2U-200m. Groups use the planar distance from the ego LiDAR origin to the box center; AP@0.5 is reported as a percentage.}
\label{tab:range_wise_results}
\centering
\footnotesize
\setlength{\tabcolsep}{2.4pt}
\begin{tabular*}{\columnwidth}{@{\extracolsep{\fill}}lcccc@{}}
\hline
& \multicolumn{4}{c}{Distance to Ego (m)} \\
\cline{2-5}
Method & 0--50 & 50--100 & 100--150 & 150--200 \\
\hline
GT boxes & 727 & 4,531 & 6,198 & 7,498 \\
\hline
Single USV & 80.99 & 76.23 & 67.35 & 51.21 \\
Late Fusion & 81.49 & 80.34 & 78.88 & 71.44 \\
Early Fusion & \textbf{85.53} & \textbf{81.72} & \textbf{82.46} & \textbf{79.59} \\
\hline
F-Cooper & 81.43 & 74.17 & 71.42 & 61.12 \\
AttFusion & 78.02 & 75.84 & 73.66 & 63.83 \\
Where2comm & 75.57 & 79.90 & 74.90 & 66.97 \\
\hline
\end{tabular*}
\end{table}

\begin{table*}[t]
\caption{Robustness of LiDAR cooperative detection under spatial pose noise and temporal delay on MarU2U-200m. AP@0.5 is reported as a percentage. $P_k$ denotes Gaussian pose noise with $\sigma_t=k$ m and $\sigma_r=k^\circ$, and $T_\Delta$ denotes a temporal delay of $\Delta$ ms.}
\label{tab:robustness_noise}
\centering
\footnotesize
\setlength{\tabcolsep}{4.0pt}
\begin{tabular*}{\textwidth}{@{\extracolsep{\fill}}lcccccccccc@{}}
\hline
Method & Clean & $P_{0.4}$ & $P_{0.8}$ & $P_{1.2}$ & $T_{200}$ & $T_{400}$ & $T_{600}$ & $P_{0.4}+T_{200}$ & $P_{0.8}+T_{400}$ & $P_{1.2}+T_{600}$ \\
\hline
Single USV & 64.11 & 64.11 & 64.11 & \textbf{64.11} & 64.11 & 64.11 & 64.11 & 64.11 & 64.11 & \textbf{64.11} \\
Late Fusion & 76.93 & 74.04 & 65.55 & 58.98 & 76.63 & 74.85 & 70.08 & 73.02 & 62.04 & 55.12 \\
Early Fusion & \textbf{81.44} & \textbf{78.35} & \textbf{68.24} & 58.01 & \textbf{80.95} & \textbf{78.23} & \textbf{72.79} & \textbf{76.93} & \textbf{64.28} & 53.48 \\
\hline
F-Cooper & 68.81 & 66.53 & 58.48 & 51.13 & 68.32 & 65.10 & 59.54 & 65.06 & 54.40 & 45.67 \\
AttFusion & 70.97 & 68.83 & 60.27 & 54.10 & 70.85 & 68.53 & 65.12 & 67.62 & 57.49 & 49.61 \\
Where2comm & 73.63 & 71.58 & 63.80 & 57.18 & 73.64 & 71.73 & 68.65 & 70.87 & 61.33 & 53.76 \\
\hline
\end{tabular*}
\end{table*}

Table~\ref{tab:scale_wise_results} reports LiDAR detection performance by vessel length. Most methods perform best on vessels between 10 and 30~m, while vessels at least 100~m long remain the most challenging group. Larger vessels are not necessarily easier to detect. For very large vessels, observing part of the hull may still leave the box center and full extent difficult to estimate. Early Fusion achieves the highest AP@0.5 in all four size groups, improving detection of the largest vessels from 30.51 to 54.88 over Single USV. These results highlight the value of complementary viewpoints and identify detection of very large vessels as a distinct challenge in maritime cooperative perception.

Table~\ref{tab:range_wise_results} shows that cooperative gains become more pronounced at longer ranges. Early Fusion improves AP@0.5 over Single USV by 4.54 percentage points at 0--50~m and 28.37 percentage points at 150--200~m. In the farthest interval, all cooperative methods outperform Single USV, with Early Fusion achieving the highest AP@0.5 of 79.59. A vessel far from the ego USV may be observed at a shorter distance or from a less occluded viewpoint by another USV. Such observations can supplement sparse or incomplete ego measurements. These results show that cooperation can mitigate the decline in detection performance with increasing distance from the ego USV.

\subsection{Robustness to Pose Noise and Temporal Delay}

We evaluate LiDAR cooperative detection under pose noise and temporal delay without retraining. Zero-mean Gaussian noise is applied to non-ego USV poses, with paired translation and rotation standard deviations of 0.4~m/$0.4^\circ$, 0.8~m/$0.8^\circ$, and 1.2~m/$1.2^\circ$. Temporal delays of 200, 400, and 600~ms use delayed non-ego observations and poses, while retaining the current ego observation and ground truth. We also evaluate combinations of the corresponding pose-noise and delay levels. All conditions, including the clean reference, use the same frames with sufficient delay history.

Table~\ref{tab:robustness_noise} shows that pose noise causes greater degradation than temporal delay under the tested settings. Early Fusion achieves the highest AP under clean observations and all temporal delay settings, while Late Fusion ranks highest among cooperative methods under the strongest pose noise and combined perturbation. Spatial alignment is particularly important over long sensing ranges: a yaw error of $1.2^\circ$ produces approximately 4.2~m of lateral displacement at 200~m. Such displacement can disrupt the alignment of vessel observations across USVs. Under the strongest pose noise or combined perturbation, all cooperative methods fall below the Single USV baseline. These results highlight the importance of accurate relative poses for maintaining the benefits of maritime cooperative perception.

\section{Transfer to Real Maritime Data}

\begin{table}[t]
\caption{Real-world LiDAR transfer on the SeePerSea dataset. AP values are reported as percentages.}
\label{tab:lidar_transfer_real}
\centering
\footnotesize
\setlength{\tabcolsep}{2.0pt}
\begin{tabular*}{\columnwidth}{@{\extracolsep{\fill}}c@{\hspace{5pt}}ccc@{\hspace{7pt}}ccc@{}}
\hline
& \multicolumn{3}{c}{Random} & \multicolumn{3}{c}{MarU2U pretrain} \\
Real data & AP@0.3 & AP@0.5 & AP@0.7 & AP@0.3 & AP@0.5 & AP@0.7 \\
\hline
10\% & 26.30 & 7.41 & 0.07 & \textbf{59.75} & \textbf{38.10} & \textbf{2.36} \\
20\% & 48.65 & 30.73 & 3.46 & \textbf{74.94} & \textbf{57.27} & \textbf{10.42} \\
50\% & 63.35 & 47.74 & 17.35 & \textbf{85.27} & \textbf{69.14} & \textbf{20.32} \\
100\% & 71.20 & 55.13 & 25.46 & \textbf{88.67} & \textbf{76.60} & \textbf{38.12} \\
\hline
\end{tabular*}
\end{table}

\begin{table}[t]
\caption{Image-based transfer to real maritime datasets (AP@0.5/AP@0.5:0.95).}
\label{tab:rgb_transfer_real}
\centering
\footnotesize
\renewcommand{\arraystretch}{1.03}
\begin{tabular*}{\columnwidth}{@{\extracolsep{\fill}}llcccc@{}}
\hline
Dataset & Init. & 1\% & 5\% & 10\% & 100\% \\
\hline
SMD & Scratch & -- & -- & -- & \textbf{99.4}/94.2 \\
& COCO & 95.9/70.7 & 98.4/84.1 & 99.0/88.1 & \textbf{99.4}/95.3 \\
& MarU2U & \textbf{96.6/75.2} & \textbf{98.7/86.5} & \textbf{99.1/89.8} & \textbf{99.4/95.6} \\
\hline
MarDCT & Scratch & -- & -- & -- & \textbf{99.5}/97.2 \\
& COCO & 95.3/77.3 & \textbf{99.5}/88.0 & \textbf{99.5}/89.9 & \textbf{99.5}/97.2 \\
& MarU2U & \textbf{95.8/80.3} & \textbf{99.5/89.7} & \textbf{99.5/91.9} & \textbf{99.5/97.7} \\
\hline
SeaShips & Scratch & -- & -- & -- & 99.3/81.7 \\
& COCO & \textbf{84.4/46.6} & \textbf{95.7}/64.6 & 96.9/70.9 & 99.3/\textbf{85.2} \\
& MarU2U & 72.4/32.8 & 95.5/\textbf{65.8} & \textbf{97.1/71.0} & \textbf{99.4}/85.0 \\
\hline
\end{tabular*}
\end{table}
Real maritime datasets for multi-USV cooperative perception remain scarce. We therefore use existing single-platform datasets for transfer evaluation, including LiDAR observations from SeePerSea~\cite{SeePerSea} and images from SMD~\cite{SMD}, MarDCT~\cite{mardct}, and SeaShips~\cite{SeaShips}. These experiments assess whether data generated by MMUSV-Sim provide useful pretraining for real maritime detection.

\subsection{LiDAR Transfer}
We train the same detector architecture on four nested subsets of the SeePerSea training set, using either random initialization or a checkpoint pretrained on MarU2U-200m. Pretraining matches the real-data setup, with $0.2 \times 0.2$~m pillars and a horizontal range of $[-100,100]$~m along each axis. Both settings use identical real training samples and training configurations. Checkpoints are selected using the lowest validation loss. All models are evaluated on the official SeePerSea validation split, using BEV AP at IoU thresholds of 0.3, 0.5, and 0.7.

As shown in Table~\ref{tab:lidar_transfer_real}, MarU2U-200m pretraining improves AP at every IoU threshold and real-data fraction. With 10\% of the real training data, pretraining raises AP@0.5 from 7.41 to 38.10. With 20\% of the data, it achieves 57.27 AP@0.5, exceeding the 55.13 obtained by random initialization with the full training set. With all real training data, pretraining further raises AP@0.5 from 55.13 to 76.60 and AP@0.7 from 25.46 to 38.12. These results demonstrate that MMUSV-Sim provides transferable LiDAR training data, improving both real-data efficiency and detection accuracy.

\subsection{RGB Transfer}
We first train YOLOv8~\cite{yolov8} from random initialization on the MarU2U-200m images and annotations. The resulting detector achieves 86.9 AP@0.5 and 67.9 AP@0.5:0.95 on the MarU2U-200m image test split. Its weights are then used to initialize fine-tuning on different fractions of the training data from the three real image datasets. Pretraining on COCO~\cite{COCO}, a large-scale dataset of real images, serves as a strong generic baseline. Training from scratch on the full real training set is also included as a reference. All models use the same split of each real dataset for evaluation, with 2D detection performance measured by AP@0.5 and AP@0.5:0.95.

As shown in Table~\ref{tab:rgb_transfer_real}, MarU2U-200m pretraining matches or exceeds COCO pretraining on SMD and MarDCT at every real-data fraction. The improvements are particularly evident in AP@0.5:0.95 when real annotations are limited. On SeaShips, COCO performs better with 1\% of the training data, while MarU2U-200m becomes competitive at higher fractions. These results demonstrate the value of simulated maritime images as a pretraining source for real vessel detection.

\section{Conclusions}
This paper presented MMUSV-Sim, a perception-oriented maritime simulation and data-generation platform for multi-USV cooperative perception. Using the platform, we constructed MarU2U-200m and established a cooperative BEV detection benchmark with LiDAR and RGB--LiDAR inputs. Experiments show that complementary USV observations improve detection, particularly at long range, and that RGB information can provide additional gains. LiDAR and RGB transfer experiments further demonstrate the value of synthetic pretraining for real maritime detection.  
Future extensions could incorporate vessel hydrodynamics and wave--vessel interactions. We plan to release the simulator executable, data acquisition code, MarU2U-200m dataset, and benchmark implementation.

\bibliographystyle{IEEEtran}
\bibliography{references}

\end{document}